# From Pixels to Explanations: Interpretable Diabetic Retinopathy Grading with CNN-Transformer Ensembles, Visual Explainability and Vision-Language Models

Pir Bakhsh Khokhar[a,i,1], Carmine Gravino[a], Fabio Palomba[a], Sule Yildirim Yayilgan[b] and Sarang Shaikh[b]

[a]*Department of Informatics, University of Salerno, Via Giovanni Paolo II, 132, Fisciano, 84084, Salerno, Italy*
[b]*Department of Information Security and Communication Technology (IIK), Norwegian University of Science and Technology, Gjøvik, 2815, Norway*

**ABSTRACT**

The quality of diabetic retinopathy (DR) screening relies on the ability to correctly grade severity; however, many deep-learning (DL) classifiers cannot be easily interpreted in the clinical context. This study presents a methodology that combines strong discriminative models with multimodal explanations, converting retinal pixels into clinically interpretable outputs. Using the APTOS 2019 benchmark, we evaluated six representative CNN- and transformer-based backbones under a controlled protocol with stratified five-fold cross-validation. We then compared ensembling strategies (hard voting, weighted soft voting, stacking) and investigated a hybrid class-level fusion variant to exploit grade-specific advantages. For interpretability, we produced Grad-CAM++ visual attribution maps and short textual rationales using vision–language models (VLMs) conditioned on the fundus image and classifier outputs under conservative prompting constraints. Modern CNN backbones (ResNet-50 and ConvNeXt-Tiny) provided the strongest single-model baselines, with cross-validated QWK up to 0.919 and 0.914, respectively. Ensembling improved ordinal agreement, and weighted soft voting was the most consistent across folds (QWK 0.934 , 0.017). Hybrid class-level fusion was competitive but did not yield a statistically reliable improvement over standard fusion in paired fold comparisons (Holm-adjusted $p \geq 1.000$). For explanation quality, Grad-CAM++ offered plausible but coarse localization, and VLM rationales were generally grade-consistent. Quantitatively, VLM variants showed a trade-off between clinical completeness and template-level semantic similarity (coverage 0.700 vs. BERTScore 0.072), while image–text alignment was comparable (CLIPScore ù 0.34). Overall, the study shows that DR grading performance and multimodal explainability can be evaluated jointly with explicit protocols, motivating external validation and expert assessment.

## 1. Introduction

Diabetic retinopathy (DR) is a progressive microvascular complication of diabetes that causes vision loss. The global diabetes epidemic is increasing the scale of the problem: the International Diabetes Federation (IDF) estimates that 589 million adults (aged 2079 years) will have diabetes by 2024, and this number will increase to 853 million by 2050. The same IDF report notes that diabetes caused 3.4 million deaths in 2024 and at least USD 1 trillion in health expenditures(International Diabetes Federation (IDF), 2024a,b). The prevalence of DR has already reached a high level and is expected to rise further in the future. The global incidence of DR is estimated to be 103 million by 2045, with projections increasing to 160.5 million. The same projections also indicate that 44.8 million people will have *vision-threatening DR (VTDR)* and 28.6 million will have *diabetic macular edema (DME)* by 2045 Rai et al. (2025); Chong et al. (2024). Public health agencies also emphasize this burden: the World Health Organization (WHO) notes that diabetes-related eye diseases, including DR, cause significant vision impairment and recommends screening and timely treatment as key prevention strategies; with an estimated 950,000 people affected World Health Organization (WHO) (2026); World Health Organization Regional Office for Europe (2026), DR is described as the leading cause of vision impairment/blindness in the WHO European Region alone.

Clinically, management relies on severity grading because treatment urgency and follow-up intervals

---

[i]Corresponding author and Principal Corresponding author.

p.khokhar@studenti.unisa.it (P.B. Khokhar); gravino@unisa.it (C. Gravino); fpalomba@unisa.it (F. Palomba); sule.yildirim@ntnu.no (S.Y. Yayilgan); sarang.shaikh@ntnu.no (S. Shaikh)

https://rubrica.unisa.it/persone?matricola=066182 (P.B. Khokhar); https://docenti.unisa.it/004724/home (C. Gravino); https://docenti.unisa.it/027888/home (F. Palomba); https://www.ntnu.edu/employees/sule.yildirim (S.Y. Yayilgan); https://www.ntnu.edu/employees/sarang.shaikh (S. Shaikh)

ORCID(s):

[1]This is the first author footnote, also applicable to the third author.

depend on the disease stage. In many screening programs, fundus photographs are graded into five ordinal categories (No DR, Mild, Moderate, Severe, and Proliferative DR), reflecting an increasing burden of microaneurysms, hemorrhages, and advanced neovascular changes. The five levels are also used in widely adopted public benchmarks, such as APTOS 2019 Karthik et al. (2019), which provides a standardized dataset and an ordinal-aware evaluation protocol.

Deep learning has become the dominant approach for grading DR from retinal fundus images over the last decade, owing to improved representation learning and large datasets. Despite high predictive performance, most models are difficult to interpret. A predicted grade in clinical workflows is rarely sufficient on its own; clinicians and screening programs must understand whether the model relies on plausible retinal evidence (e.g., lesion patterns) or confounding cues (e.g., imaging artifacts, illumination changes, or camera-specific biases). This is reflected in studies that focus on strong assessments and ordinal metrics that have clinical significance, such as quadratic weighted kappa (QWK). In cases in which there is a significant difference between the predicted and true grades, it leads to stronger penalization of disagreements (Chilukoti et al., 2024; Akhtar et al., 2025).

In practice, grading is difficult because of subtle class boundaries and imbalanced data: mild DR is underrepresented in many screening datasets, whereas advanced stages are less prevalent owing to treat-ment and referral processes. This encourages models to optimize overall accuracy while underperforming on minority grades, which are often the most clinically significant.

Ensemble learning has remained attractive for DR because complementary decision boundaries can reduce grade-specific failure modes. A recent study on DR grading using transfer learning and ensembles demonstrated that combining multiple backbones can improve ordinal agreement across standard benchmarks (Chilukoti et al., 2024; Akhtar et al., 2025). In addition, modern architectures encompass both convolutional networks and vision transformers, motivating hybrid combinations that can combine local texture sensitivity (CNNs) with stronger global context modeling (transformers) in one pipeline (Boulaabi et al., 2025; Aburass et al., 2025).

In such systems, explainability is commonly added using post hoc visual attribution methods, such as Grad-CAM and its variants. Grad-CAM provides an intuitive heatmap overlay showing where the model "looked" to support a decision by backpropagating gradients to a chosen convolutional layer. When DR is performed, heatmaps are often used to determine whether the model focused on lesion-rich regions. However, recent interpretability analyses on DR classification have shown that different attribution methods can yield inconsistent or conflicting explanations for the same image, and that explanation behavior can be strongly affected by the architecture and implementation details (Zhang et al., 2025a). A visually appealing heatmap does not necessarily translate into a clinically usable rationale for DR lesions, and practitioners often have difficulty converting diffuse saliency into a concrete interpretation of DR lesions because of their small size and spatial dispersion. These observations highlight three important limitations of explainability in current DR grading systems.

- First, heatmap-based explanations may vary substantially across attribution methods, reducing their consistency and reliability.

- Second, their behavior can depend strongly on the underlying architecture and implementation choices, making interpretation less stable across models.

- Third, even when visually plausible, saliency maps often remain too diffuse to provide a concrete and clinician-readable rationale for small and spatially scattered DR lesions.

As a result of these limitations, multimodal explanations are becoming more common, bringing them closer to clinical reporting. Vision-language models (VLMs) can translate visual evidence into short textual explanations, potentially bridging the gap between model internals and clinician-facing interpretations. Recent reviews of medical vision-language foundation models have noted rapid progress in multimodal instruction following and biomedical image understanding. However, medical VLMs must be carefully evaluated and domain-specifically grounded. In particular, robustness remains a key issue: VLM results are susceptible to artifacts and changes in distributions, and systematic studies have demonstrated that medical VLMs can be adversely affected by realistic imaging perturbations and do not perform well in real-world images (and other images) (Cheng et al., 2025). These findings underline that generating text is insufficient; explanation quality must be measured, and reliability risks must be explicitly addressed.

Motivated by these limitations, our work moves beyond heatmap-only explainability by combining visual attribution with vision-language-based textual rationales, with the goal of providing explanations that are not only visually grounded but also more clinically interpretable and explicitly evaluable.

Motivated by these trends, we propose a unified DR grading and explainability pipeline that moves from *pixels to explanations*. We first build discrimi-native graders by benchmarking representative CNN and transformer backbones on APTOS 2019 and then examine multiple ensembling strategies to improve robustness under subtle grade boundaries and class imbalance. We generate visual explanations using Grad-CAM++ overlays and condition a medical vision language model to produce concise textual rationales, using the predicted grade probabilities (and, where applicable, localization cues) as structured context in the prompt (Selvaraju et al., 2017; Li et al., 2023). Finally, we treat explanation generation as an evaluable output: beyond qualitative examples, we assess textual explanations in terms of semantic fidelity to grade-specific references, visual grounding with the corresponding image, and coverage of clinically relevant retinal descriptors.

Accordingly, the study unfolds in three connected stages: it first establishes a baseline across backbone architectures, then examines whether ensembling can improve robustness, and finally investigates whether stronger grading performance can be accompanied by clinically meaningful multimodal explanations. Our main contributions are:

**Backbone Benchmarking**

We benchmark six CNN and transformer backbones on APTOS 2019 for five-class DR grading using standard metrics and QWK. This comparison provides a controlled assessment of how differ-ent architectural families perform under the same training, validation, and evaluation protocol.

**Hybrid Class-Level Fusion**

We evaluate multiple ensembling strategies (hard voting, weighted soft voting, stacking) and investigate a *hybrid class-level* fusion variant for grade-wise specialization.

**Multimodal Explainability**

We generate multimodal explanations by combining Grad-CAM localization with VLM-based textual rationales.

**Explanation Quality Evaluation**

We evaluate explanations in terms of semantic similarity to reference rationales, image–text alignment, and coverage of clinically relevant retinal descriptors.

The remainder of this paper is organized as follows. **Section 2** discusses recent progress in DR grading, ensemble learning, and explainability, with an emphasis on multimodal (vision-language) approaches. **Section 3** presents the materials, study design, and experimental setup, including the dataset and cross-validation protocol, preprocessing and augmentation, backbone architectures, training objective and optimization, ensembling strategies (including the proposed hybrid class-level scheme), the explainability pipeline (Grad-CAM and VLM-based rationales), evaluation metrics, and implementation details related to reproducibility. **Section 4** reports quantitative results for individual backbones and ensembles, followed by qualitative and quantitative evaluations of graphical and textual explanations. **Section 5** discusses implications, failure modes, robustness considerations, and limitations. Finally, **Section 6** concludes the paper and outlines directions for external validation and clinically grounded explanation assessment.

## 2. Related Work

For years, researchers have worked on automating the detection of diabetic retinopathy using color fundus images. This push has come from the growing demand for wider screening that does not depend solely on human experts. Standardized public datasets have shaped much of this development, offering a common ground for defining tasks and fairly measuring results. Owing to these resources, assessing how well different systems classify disease stages has become more reliable. They also allow clearer contrasts between various models and methods for training them (Chilukoti et al., 2024). Recently, greater attention has been paid to building systems resilient to differences in image clarity and uneven case distributions. At the same time, accuracy in matching clinical assessments remains a key measure of success (Chilukoti et al., 2024).

Although older, convolutional neural networks hold up well because of their built-in assumptions about nearby pixel relationships and subtle anomaly detection. Because medical image sets often lack size or balance, practitioners continue to rely on pretrained models from vast photo collections to speed training and boost performance. Recently, efforts targeting diabetic retinopathy have shifted toward reshaping model structures, for example, using attention layers, to reflect varied lesion appearances and navigate complex classification edges. For example, CRA-Net introduces transformer-guided category-relation attention to model long-range dependencies while improving within-class representation (Zang and Ma, 2024). In addition to such architectural contributions, several recent grading pipelines have reported strong ordinal agreement using modern CNN families and task-specific refinements, including EfficientNet-based designs with attention or gating (Dharrao et al., 2025) and transfer-learning frameworks designed explicitly for stage grading (Akhtar et al., 2025). Comparative evidence also suggests that modern model families can differ meaningfully in their error profiles and decision behavior. For instance, vision transformers have been reported to outperform CNN baselines for referable

DR detection, highlighting the potential value of global context modeling in fundus-based analysis (Goh et al., 2024).

In parallel, vision transformers and hybrid CNN Transformer architectures have become increasingly prominent in fundus-based DR research. The main motivation is to combine CNN sensitivity to fine-grained local texture with the transformers capacity to model global structure and long-range dependencies. Recent studies have proposed transformer-based DR systems and hybrid designs aimed at improving classification performance and computational efficiency (Huang et al., 2025). Hybridization has also been explored in practical grading pipelines. For example, hybrid EfficientNet–Swin designs have been evaluated to combine local lesion sensitivity with contextual modeling (Xu et al., 2024). Comparative work has also moved beyond reporting accuracy alone by examining how CNNs, ViTs, and hybrid models behave in practice, including their error patterns and explanation stability. These studies suggest that architectural choice can affect not only grading performance but also the type and consistency of the explanations produced (Zhang et al., 2025b). Overall, the literature suggests that hybridization can be beneficial; however, the magnitude of gains varies across experimental settings, reinforcing the need for careful benchmarking and grade-wise analysis rather than reliance on a single best backbone.

In addition to single models, ensemble learning remains a common strategy for improving robust-ness under label noise, acquisition variability, and subtle boundaries between adjacent grades. Standard ensemble approaches include hard voting, probabil-ity averaging (soft voting), and stacking-based meta-learning. Recent work reports that combining multiple models can improve severity-grading performance compared with individual backbones, supporting the view that ensembles can reduce grade-specific failure modes (Chilukoti et al., 2024). Related directions have also explored alternative fusion strategies beyond classic ensembling, such as representation-fusion approaches that combine complementary feature signals (Mohsen et al., 2025) and multi-scale attention-based grading networks designed to improve fine-grained staging (Sun et al., 2025). This broader body of work aligns with guidance in medical imaging that highlights pitfalls in metric interpretation and encourages transparent reporting across complementary measures matched to the clinical objective (Kocak et al., 2025).

Explainability has also become increasingly important because clinical stakeholders typically require not only a predicted grade but also evidence supporting the decision. Visual attribution methods are the most common, with Grad-CAM and its variants frequently used to highlight influential regions (Selvaraju et al., 2017). However, DR lesions can be small and spatially dispersed, and post-hoc heatmaps are not guaranteed to be stable, specific, or clinically faithful. A DR-focused interpretability study comparing multiple explanation techniques reported substantial variability across methods and architectures, including sensitivity to modeling choices and potential emphasis on non-lesion structures or artefacts (Zhang et al., 2025a). These observations motivate explanation pipelines that go beyond heatmaps alone and provide more explicit, human-readable rationales while maintaining a clear evaluation protocol.

More recently, VLMs have enabled multimodal explanations by generating textual rationales conditioned on images and auxiliary signals, such as predicted probabilities or localization cues. LLaVA-Med demonstrates that instruction-tuned VLMs can produce clinically styled responses when adapted to biomedical image–text data (Li et al., 2023). Nevertheless, reliability concerns remain central, as medical VLM outputs can degrade under common image artifacts, raising practical concerns for deployment (Cheng et al., 2025). The evaluation of ophthalmology-focused multimodal large language models is also becoming more systematic. Recent benchmarking studies of fundus interpretation have shown that these models still struggle with several basic ophthalmic tasks, highlighting the need for domain-specific training and more rigorous assessment frameworks (Wei et al., 2025). In the same direction, specialized training curricula have been proposed to improve vision–language model performance in retinal image analysis, suggesting a broader shift from generic prompting strategies toward ophthalmology-adapted multimodal learning (Holland et al., 2025). Meanwhile, recent retinal diagnostic pipelines have begun to incorporate vision–language components directly into end-to-end systems designed for transparency, reflecting growing interest in explanation generation as a core output rather than a secondary illustrative feature (La Rosa et al., 2026).

These reliability concerns have strengthened the case for a quantitative evaluation of generated explanations instead of relying only on selective qualitative examples. Metrics such as BERTScore enable semantic comparisons between generated and reference texts, whereas CLIPScore offers a reference-free measure of image–text consistency (Zhang et al., 2020; Hessel et al., 2021). In parallel, community benchmarks in medical captioning and interpretable report generation have increasingly emphasized structured evaluation schemes, including concept-level assessments and explainability-oriented criteria (Damm et al., 2025). Taken together, these developments support the need for pipelines that combine accurate grading models with multimodal explanatory outputs while evaluating those explanations explicitly, rather than treating them as supplementary illustrations.

To clarify the motivation for our study and its position within the existing literature, we summarize below the main limitations of prior work and the specific contributions of our approach.

> ⚠ **Grade-wise instability**
>
> Many DR grading systems rely on a single backbone model, which often results in uneven performance across disease stages and continued confusion between neighboring grades.

> ⚠ **Heatmap-only explanations**
>
> In many studies, explainability is restricted to saliency maps or Grad-CAM visualizations. Although useful, these outputs can be diffuse, sensitive to the explanation method, and difficult to interpret for clinician-oriented reasoning.

> ⚠ **Limited explanation evaluation**
>
> Explanation quality is often assessed only qualitatively or with metrics that do not adequately reflect clinical faithfulness, relevance, and completeness.

Our study makes three main contributions. First, we conduct a controlled benchmark of representative CNN and transformer backbones for multi-class DR grading using a unified training pipeline and a consistent evaluation protocol. Second, we propose a hybrid class-level ensembling strategy that leverages grade-wise complementarity across heterogeneous architectures to reduce class-specific error patterns and improve overall robustness. Third, we develop a multimodal explainability pipeline that combines Grad-CAM-based visual evidence localization with VLM-generated textual rationales, and we quantitatively assess explanation quality using BERTScore, CLIPScore, and a domain-driven coverage metric. The next section describes the dataset and preprocessing steps, the backbone and ensembling configurations, the Grad-CAM and VLM prompting framework, and the evaluation protocols used for both grading performance and explanation quality.

## 3. Materials and Methods

This section presents the dataset, preprocessing pipeline, model backbones, ensembling strategies, explainability framework, and evaluation protocols used in this study. It includes the proposed hybrid class-level ensembling scheme, Grad-CAM-based visual evidence localization, VLM-generated textual rationales, and the evaluation of both grading and explanation quality.

**Research Questions**

To structure the experimental design around clear objectives, we addressed the following research questions:

> 🔍 **RQ1**
>
> How do representative CNN and transformer backbones compare for five-class DR grading under a unified training and evaluation protocol?

> 🔍 **RQ2**
>
> Do standard ensembling strategies (hard voting, weighted soft voting, stacking) and a hybrid class-level ensemble improve grade-wise robustness relative to single models?

> 🔍 **RQ3**
>
> To what extent can multimodal explanations that combine Grad-CAM localization with VLM generated text provide clinician-readable rationales, and how reliably can explanation quality be quantified using BERTScore, CLIPScore, and coverage?

These three research questions follow a clear and logical progression. *RQ1* first establishes a baseline by comparing representative CNN and transformer backbones under the same training and evaluation settings, allowing us to see how architectural choices influence ordinal DR grading. Building on this, *RQ2* investigates whether combining these models can reduce their individual weaknesses, especially in the presence of class imbalance and confusion between adjacent grades. *RQ3* then takes the analysis a step further by moving beyond prediction accuracy to interpretability, asking whether the final grading pipeline can also provide clinically meaningful multimodal explanations. In this way, the study progresses from backbone comparison to robustness through ensembling, and finally to explanation quality.

**Dataset and Severity Labels**

We conducted five-class DR severity grading using the APTOS 2019 fundus image benchmark, which contains 3,662 labeled images annotated into five ordinal stages: no DR, mild, moderate, severe, and proliferative DR (Karthik et al., 2019). The images were collected under heterogeneous acquisition conditions (different cameras, illumination, and fields of view), which introduced substantial variability in color balance, contrast, blur, and background artifacts that could confound automated grading (Karthik et al., 2019). As shown in Figure 1, the class distribution was markedly imbalanced, with early stages dominating and severe classes occurring far less frequently; we explicitly accounted for this imbalance during training and evaluation.

APTOS 2019 is well suited to the aims of this study for several reasons. First, its five-class ordinal label structure closely reflects the clinical grading setting addressed by our research questions. Second, the strong class imbalance makes it a useful benchmark for testing whether different backbone architectures and ensemble strategies can maintain reliability on underrepresented severity grades rather than performing well only on the majority classes. Third, the variability in acquisition conditions makes the dataset valuable not only for evaluating predictive performance but also for assessing how explainability methods behave when image quality, contrast, and background characteristics change. For these reasons, APTOS 2019 was selected not only because it is widely used in prior work but also because its properties directly support the evaluation of baseline performance, robustness through ensembling, and multimodal explanation quality.

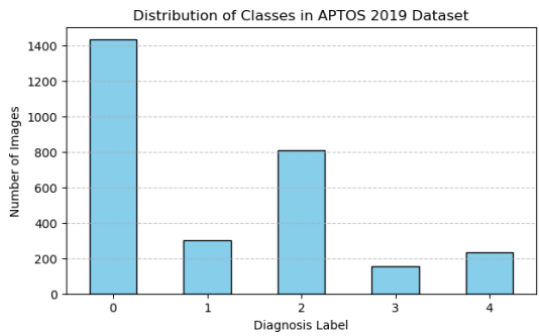

Figure 1: Distribution of DR grades (0–4) in APTOS 2019. The imbalance across severity levels motivates imbalance-aware training and the use of complementary evaluation metrics (e.g., macro-F1 and QWK).

**Cross-Validation Protocol**

We employed stratified five-fold cross-validation to obtain performance estimates that are less sensitive to a single split and remain representative under class imbalance. The dataset was partitioned into five folds using stratified sampling on the DR grade labels, ensuring that the class distribution was approximately preserved within each fold. For each cross-validation run, four folds were used for training, and the remaining fold was used for validation; this was repeated five times so that each fold served as the validation set exactly once. Fold assignment was generated with shuffling enabled and a fixed random seed (random_state = 42) to ensure reproducibility.

To minimize the risk of information leakage, model selection and hyperparameter tuning were performed separately for each training-validation split. Specifically, checkpoint selection is based only on the validation performance for the corresponding fold, and the selected checkpoint is then used to compute the reported metrics for that fold. The final results are reported as the mean and standard deviation across folds. This protocol provides a controlled and repeatable evaluation framework, supports fair comparisons across backbones and ensembles, and enables grade-wise performance assessment in the presence of imbalanced class distributions.

**Image Preprocessing and Augmentation**

Fundus photographs exhibit substantial variability owing to acquisition conditions, including illumination changes, contrast variation, focus or blur, and sensor-specific artifacts, all of which can confound automated grading. To ensure a controlled comparison across backbones, we applied the same preprocessing pipeline to all models. The images were resized to a fixed resolution of $224 \times 224$ pixels, consistent with the standard input size commonly adopted by the selected pretrained architectures, including ResNet, ConvNeXt, DeiT, and Swin Transformer variants (He et al., 2016; Liu et al., 2022; Touvron et al., 2021; Liu et al., 2021). Local contrast was enhanced using contrast-limited adaptive histogram equalization (CLAHE), a technique commonly used in retinal image preprocessing to improve local contrast and lesion visibility (Imran et al., 2019), with a clip limit of 2.0 and an $8 \times 8$ tile grid. To avoid introducing a systematic mismatch between training and evaluation, CLAHE was applied stochastically during training ($p$ = 0.5) and deterministically during validation ($p$ = 1.0).

To improve generalization and reduce sensitivity to acquisition variability, we employed on-the-fly data augmentation during training. Geometric augmentations included random horizontal and vertical flips ($p$ = 0.5 each), random rotations by multiples of 90 ($p$ = 0.5), and mild affine perturbations (translation up to 5%, scaling up to 5%, and rotation up to 30; $p$ = 0.7). Photometric augmentation was implemented via random brightness and contrast adjustments up to ,0.2 ($p$ = 0.7). In addition, to promote robustness to localized occlusions and spurious artifacts, we applied coarse dropout by removing up to eight rectangular regions per image.

Finally, the images are normalized using ImageNet channel statistics (mean = [0.485, 0.456, 0.406], standard deviation = [0.229, 0.224, 0.225]) and converted into PyTorch tensors. The validation pipeline includes resizing, deterministic CLAHE, and normalization; however, geometric and photometric augmentations are excluded to guarantee a stable and comparable evaluation setup.

**Backbone Models**

We benchmark six backbone architectures that are widely used in recent DR and medical imaging classification literature and represent complementary inductive biases (Chilukoti et al., 2024; Zhang et al., 2025a; Goh et al., 2024). The selected set includes established convolutional baselines with strong local feature sensitivity (ResNet-50 (He et al., 2016), DenseNet-121 (Huang et al., 2017)), parameter-efficient convolutional scaling (EfficientNet (Tan and Le, 2019)), a modernized convolutional design (ConvNeXt-Tiny (Liu et al., 2022)), and transformer families that capture broader context via attention, including a data-efficient ViT variant (DeiT-Small (Touvron et al., 2021)) and a hierarchical window-based transformer (Swin Transformer (Liu et al., 2021)). For each backbone, we initialize the weights from ImageNet pretraining when available and replace the original classification head with a five-class output layer to match the DR grading task. All backbones are trained using the same input resolution, preprocessing and augmentation pipeline, data splitting protocol, loss function, and model-selection criterion to enable a controlled comparison. Unless otherwise stated, the final prediction is obtained from the softmax-normalized logits produced by the task-specific classification head.

The backbone architectures were chosen to capture potentially complementary strengths that are relevant to the goals of this study. CNN-based models are generally well suited for detecting fine-grained and localized lesion patterns, whereas transformer-based models may be better at capturing broader retinal contexts and longer-range structural relationships. Comparing these two model families under the same training conditions therefore helps to address *RQ1*, and also provides a basis for examining in *RQ2* whether their differences can be effectively exploited through ensembling.

### Training Objective and Model Selection

To address class imbalance and reduce overconfident predictions, we trained all backbones using a class-weighted cross-entropy objective with label smoothing and focal reweighting. Label smoothing was applied with a smoothing factor $\epsilon = 0.1$, and focal modulation used a focusing parameter $\mu = 2.0$ to upweight hard (low-confidence or misclassified) samples. Class weights were computed from the training labels in each fold using a balanced weighting scheme and were applied consistently during optimization.

To ensure a fair comparison across architectures, all backbones were trained under the same training protocol within each cross-validation fold. Model selection was performed strictly on the validation fold: checkpoints were evaluated on the fold-specific validation split, and the best checkpoint was selected according to the predefined validation criterion. The selected checkpoint was then used to compute the reported metrics for that fold, avoiding the use of evaluation data during training or model selection.

### Ensembling Strategies

To improve robustness under class imbalance and reduce adjacent-grade errors, we evaluated multiple standard ensembling schemes and a proposed class-level variant. For all ensembles, we used the same set of base models and operated on their predicted class probabilities and labels. The ensemble methods were selected to represent increasingly flexible ways of combining complementary models, an approach that has been widely adopted to mitigate model-specific error patterns and improve robustness in medical image classification (Chilukoti et al., 2024; Akhtar et al., 2025). Hard voting serves as a simple label-level baseline, weighted soft voting examines whether combining class probabilities can produce more stable predictions in the presence of ordinal ambiguity and class imbalance, and stacking tests whether a learned meta-model can improve fixed fusion rules (Kocak et al., 2025). The proposed hybrid class-level approach further explores whether different models are more reliable for specific DR grades, which is particularly relevant when adjacent classes are difficult to distinguish and minority severity stages are less represented (Mohsen et al., 2025; Sun et al., 2025).

*Hard Voting:* Each base model outputs a class label, and the final prediction is obtained by majority voting. In the event of a tie, we select the tied class with the highest mean confidence across the models involved in the tie. This tie-breaking rule is fixed and applied consistently across all experiments.

*Weighted Soft Voting:* We combine models at the probability level by averaging their predicted class-probability vectors using non-negative weights. The weights are estimated on the validation split using a predefined criterion (e.g., macro-F1 or QWK) and are then fixed during evaluation. This approach allows stronger base models to contribute more to the final prediction while maintaining a transparent, deterministic fusion rule at test time.

*Stacking:* We implement stacking as a two-level scheme in which a multinomial logistic regression meta-learner is trained to combine base model outputs. The meta-learner input is the concatenation of base model probabilities (or logits) for each class. To avoid information leakage, the meta-learner is trained only on validation predictions generated from base models not trained on those samples (i.e., fold-consistent out-of-sample predictions). The learned meta-learner is then applied to the held-out evaluation data.

*Hybrid Class-Level Fusion:* In addition to global weighting, we evaluate a class-level fusion strategy that assigns model weights separately for each class. Class-specific weights are derived from validation behavior (e.g., class-wise recall/F1 or confusion tendencies), allowing models that are empirically stronger for particular severity stages to contribute more to those stages. Importantly, the class-wise weighting policy is estimated using only validation data and fixed for evaluation, ensuring that class-level specialization does not exploit test set information.

**Multimodal Explanation Pipeline**

To support both visual and textual interpretability, we employ a two-stage explainability pipeline that combines gradient-based localization (Grad-CAM) with vision language model (VLM) rationales. Grad-CAM and VLM-generated rationales are combined because they serve different but complementary interpretability purposes. Grad-CAM helps to show where the model is focusing in the retinal image; however, it does not explain this evidence in a manner that is easy to interpret clinically. In contrast, VLM-generated text can express the models likely reasoning in more familiar clinical language without any visual grounding, but it is more difficult to assess whether that explanation is tied to plausible image regions. Used together, these two components provide a more informative multimodal explanation by linking coarse visual evidence with concise textual justification, in line with the aim of **RQ3**.

***Visual Explanations with Grad-CAM++:*** We generated gradient-based class activation maps (Grad-CAM) for the predicted DR class (Selvaraju et al., 2017). Grad-CAM produces a class-specific saliency map by backpropagating the prediction signal to a selected late-stage feature representation, highlighting image regions that contribute the most to the decision. The resulting map was resized to the input resolution and overlaid on the corresponding fundus photograph to provide an intuitive visualization of model attention. We computed Grad-CAM overlays for both correctly classified samples and representative failure cases across severity levels to qualitatively assess whether the model focused on plausible retinal evidence.

***VLM-Based Rationale Generation:*** To generate clinician-oriented textual rationales, we employed a vision–language model (VLM) conditioned on the fundus image and classifier outputs. We evaluated two VLM variants: a lightweight general-purpose model (llava-hf/llava-1.5-1b-hf) and a biomedical-adapted model (llava-med-v1.5-mistral-7b) (Li et al., 2023). For each image, the VLM input consisted of (i) the original RGB fundus photograph and (ii) a structured textual context containing the predicted DR grade and the corresponding class-probability vector. A fixed prompt template was used to standardize the output format (2–4 sentences) and limit stylistic variance across samples.

***Prompt Design and Constraints:*** To reduce hallucination risk and encourage image-grounded reporting, we incorporate explicit constraints in the system instructions. The VLM is instructed to (i) describe only clearly visible findings, (ii) explicitly state "not clearly visible" when uncertain, (iii) avoid inventing lesions, (iv) mention relevant image-quality limitations (e.g., blur, glare, low contrast), and (v) avoid treatment recommendations. If image-grounded findings cannot be stated, the model is instructed to output a criterion-based *generic* note using conditional language (e.g., "may/typically/often") and to label the output as **GENERIC**. This design enforces a conservative explanation style and makes uncertainty explicit.

*Primary Prompt Template:* For each image, we use the following chat-style prompt, where {grade} is the predicted DR grade name, {c} is the predicted class index, and {p_0,...,p_4} are the predicted probabilities:

---

**Primary Prompt**

SYSTEM: You are an ophthalmology reporting assistant for diabetic retinopathy grading.
SYSTEM: Describe only findings that are clearly visible. If unsure, write '**not clearly visible**'.
SYSTEM: Do not infer or invent lesions. Mention image quality limitations if present. No treatment advice.
SYSTEM: If image-grounded findings cannot be provided, output a **GENERIC** grade-criterion note using conditional language (may/typically/often) and label it **GENERIC**.
USER: <image>
Fundus photograph. Predicted grade: {grade} (class {c}); probability: [{p_0,...,p_4}].
Write 2–4 sentences using concise clinical terms only. Avoid speculation.
ASSISTANT:

---

*Structured Checklist Prompt:* To examine whether explanation quality is robust to prompt formulation rather than tied to a single wording, we also evaluated a structured prompt that asks the model to explicitly indicate the presence, absence, or uncertainty of key retinal findings, followed by a brief justification.

---

**Alternative Prompt**

SYSTEM: You are an ophthalmology reporting assistant. Only report findings that are clearly visible in the image. If unsure, write '**not clearly visible**'. Do not invent lesions. No treatment advice.
USER: <image>
Predicted grade: {grade} (class {c}).
For each item, indicate Present/Absent/Not clearly visible: microaneurysms, hemorrhages, exudates, cotton-wool spots, neovascularization, and macular involvement.
Then, write one sentence explaining why the predicted grade is consistent with the visible findings.
ASSISTANT:

---

*Decoding and Output Processing:* Text generation is performed with fixed decoding parameters (maximum new tokens, temperature, and top-*p* sampling) applied uniformly across all images and both prompts. The outputs are constrained to short rationales and are post-processed to remove control tokens (e.g., chat markers and probability strings) before computing text-based metrics.

*Logging and Output Visualization:* The generated rationales are stored together with the predicted label, probability vector, and corresponding image identifier.

When Grad-CAM overlays are available, we additionally generate tripanel visualizations (original image, Grad-CAM overlay, and textual rationale) to facilitate qualitative inspection of the alignment between the predicted grade, localized evidence, and the generated explanation.

***Reproducibility Controls:*** *We used fixed model identifiers (llava-hf/llava-1.5-1b-hf and llava-med-v1.5-mistral-7b), fixed prompt text (primary and alternative), and fixed decoding parameters for all samples. The probability vector is inserted in a consistent format with values rounded to two decimal places, and the same post-processing rules are applied prior to scoring.*

## Evaluation Metrics

*Grading Performance Metrics:* We evaluated DR severity grading using a set of complementary metrics that capture both overall correctness and class-sensitive behavior under imbalance. Overall accuracy is reported as a coarse summary of correct predictions. Precision and recall are reported to characterize false-positive and false-negative tendencies, respectively, whereas macro-F1 treats all classes equally and emphasizes performance on minority grades. Weighted F1 additionally accounts for class prevalence to reflect the expected performance under the observed label distribution. Because DR grades are ordinal, we also report quadratic weighted kappa (QWK), which measures the agreement between the predicted and reference grades while penalizing larger grade discrepancies more strongly than adjacent-grade errors (Karthik et al., 2019).

*Textual Explanation Quality Metrics:* We assessed textual rationales using three complementary criteria designed to measure semantic fidelity, image grounding, and clinical completeness.

**(i) Semantic Similarity (BERTScore).** We compute BERTScore to quantify the semantic overlap between the generated explanation and a reference text using contextual token embeddings (Zhang et al., 2020). Because APTOS does not provide human-written rationales, we adopt grade-specific reference templates that describe canonical findings for each severity level (0–4) and use them as reference targets for scoring. The templates are as follows: *Class 0:* "Normal retina with clear macula and intact vasculature. No visible microaneurysms, hemorrhages, or exudates." *Class 1:* "Mild NPDR: few microaneurysms with minimal vascular changes. No significant hemorrhages or exudates." *Class 2:* "Moderate NPDR: multiple microaneurysms and dot-blot hemorrhages with scattered hard exudates, mainly around the posterior pole." *Class 3:* "Severe NPDR: numerous hemorrhages with venous bleeding and cotton-wool spots; ischemic changes may be present near the posterior pole." *Class 4:* "PDR: neovascularization with possible preretinal/vitreous hemorrhage and widespread ischemic changes." When the VLM explicitly labels its output as *generic*, we score it against the template of the predicted grade.

**(ii) Image–Text Grounding (CLIPScore):** To quantify how well a generated rationale is aligned with the associated fundus image without requiring reference text, we compute the CLIPScore using the CLIP ViT-B/32 model (Hessel et al., 2021). Before scoring, the generated text is sanitized to remove prompt/control tokens (e.g., "USER:", "ASSISTANT:", and image placeholders), markdown artifacts, and bracketed probability strings, and it is truncated to a fixed length to satisfy CLIP token limits. The processed caption and the corresponding image are embedded using the CLIP text and image encoders, respectively. Both embeddings are $\ell_2$-normalized, and the final CLIPScore is computed as their cosine similarity. A higher CLIPScore indicates stronger image–text compatibility, providing a reference-free complement to reference-based semantic scoring.

**(iii) Clinical Completeness (Coverage):** We define a lightweight coverage metric to measure whether explanations include clinically relevant terms expected in DR reporting. Let $f$ be a predefined vocabulary of ten retinal descriptors: {*microaneurysm, hemorrhage, exudate, macula, vessel, retina, optic disc, cotton wool, lesion, neovascularization*}. For a generated explanation, we compute the fraction of terms in $f$ that appear in the lowercased text via exact substring matching. Coverage is the average across the evaluation set. This metric acts as a proxy for completeness by indicating whether clinically important concepts are mentioned in the generated explanation. Although it does not assess correctness or the appropriateness of localization, it complements BERTScore and CLIPScore by explicitly rewarding the inclusion of clinically meaningful descriptors.

## Statistical Analysis of Ensemble Performance

To assess whether the differences between the ensembling strategies are robust, we performed paired statistical comparisons across the five cross-validation folds using the primary ordinal metric (QWK) and macro-F1 as a secondary measure. For each ensemble method, we computed fold-wise scores and applied a two-sided Wilcoxon signed-rank test to compare the hybrid class-level ensemble against each baseline (hard voting, weighted soft voting, and stacking) using paired fold scores (Wilcoxon, 1992). To control the family-wise error rate across multiple comparisons, p-values were adjusted using the Holm–Bonferroni procedure (Holm, 1979). In addition to p-values, we report the mean difference and a 95% bootstrap confidence interval for the fold-wise metric differences to quantify the magnitude and uncertainty of the observed effect (Tibshirani and Efron, 1993).

**Implementation and Reproducibility**

All experiments were implemented in Python using PyTorch. Data preprocessing and augmentation were performed using the Albumentations library, a widely used framework for fast and flexible image augmentation in computer vision pipelines (Buslaev et al., 2020). Stratified five-fold cross-validation was performed using scikit-learn, with shuffling enabled and a fixed random seed (random_state = 42) to ensure reproducibility. Unless otherwise stated, all backbones were initialized from ImageNet-pretrained weights and trained under identical preprocessing, split protocol, and training objective to ensure a controlled comparison across model families and ensembling strategies.

*Cross-Validation Execution and Model Selection:* The models were trained fold-wise using four folds for training and one fold for validation. Within each fold, the hyperparameters were fixed across the backbones, and checkpoint selection was performed exclusively using the validation fold to avoid test-set leakage. Class weights were computed from the training labels in each fold using a balanced weighting scheme and applied consistently within the loss function. The reported performance metrics are aggregated across folds by reporting the mean and standard deviation.

*Training Configuration:* All models were trained using the same input resolution (224 Ì 224) and the same preprocessing/augmentation pipeline described in Section 3. Optimization was performed on the available compute device (GPU when available; otherwise, CPU). Training used a focal cross-entropy objective with label smoothing, with fixed parameters ($u$ = 2.0, smoothing factor $\epsilon$ = 0.1) across all runs. Unless explicitly stated, we used identical batch formation, epoch scheduling, and stopping criteria across backbones to ensure a fair comparison. Model outputs were stored as logits and class probabilities for subsequent ensemble construction and analysis.

*Ensemble Construction:* Ensembling strategies were constructed using base model outputs obtained from cross-validation runs. Hard voting used a majority decision over predicted labels, whereas soft voting combined predicted probabilities using validation-derived weights. For stacking, a multinomial logistic regression meta-learner was trained on validation predictions to learn an optimal combination of base models. For hybrid class-level ensembling, class-specific weights were estimated from validation behavior (e.g., class-wise performance patterns) and applied to base model probabilities at inference time.

*Explainability Computation and Logging:* For visual interpretability, Grad-CAM maps were generated for the predicted class and overlaid on the correspond-ing input images. For textual rationales, we used two VLM variants: llava-hf/llava-1.5-1b-hf and llava-med-v1.5-mistral-7b (Li et al., 2023). VLM inference was run with gradients disabled and a fixed decoding configuration (maximum new tokens, temper-ature, and top-*p* sampling) to standardize explanation length and style. For each evaluated image, we logged the predicted label, probability vector, Grad-CAM path, and generated reasoning in a structured results file. We also produced tripanel visualizations (original image, Grad-CAM overlay, and textual reasoning) for representative samples to support qualitative inspection.

*Reproducibility Controls:* To improve reproducibility, the data split seed was fixed, and all backbones were trained under the same preprocessing pipeline and objective. The metrics are reported consistently across folds, and all outputs required for re-evaluation (predictions, probability vectors, and generated explanations) are stored to enable independent verification of grading and explanation scores.

## 4. Results

All results are reported under the stratified five-fold cross-validation protocol described in Section 3. For each backbone and ensembling strategy, we compute fold-specific performance on the held-out validation fold and summarize the results as the mean , standard deviation across the five folds. Given the ordinal nature of DR grading, QWK is treated as the primary metric, whereas macro-F1 and weighted-F1 are reported to capture class-balanced and prevalence-weighted behavior, respectively. For explainability, Grad-CAM(++) overlays and VLM-generated rationales are produced for evaluation samples in each fold, and text quality is summarized using BERTScore (template-referenced), CLIPScore (reference-free image–text alignment), and a vocabulary-based coverage metric, which is also reported as the mean , standard deviation across folds.

### RQ1: Overall Performance Across Backbones

To answer RQ1, we benchmark six representative backbones spanning convolutional and transformer families under an identical experimental protocol, including the same preprocessing and augmentation pipeline, stratified cross-validation procedure, and training objectives. This design isolates the effect of backbone choice and enables a direct comparison of how

Table 1: Single-backbone performance under five-fold cross-validation (mean , SD). Higher is better.

| Backbone | Macro-F1 | Weighted-F1 | QWK | Macro-AUC |
| --- | --- | --- | --- | --- |
| ResNet-50 | 0.728 , 0.063 | 0.858 , 0.030 | 0.919 , 0.014 | 0.962 , 0.011 |
| ConvNeXt-Tiny | 0.732 , 0.045 | 0.848 , 0.015 | 0.914 , 0.026 | 0.956 , 0.012 |
| Swin Transformer | 0.705 , 0.056 | 0.835 , 0.017 | 0.916 , 0.006 | 0.954 , 0.009 |
| DeiT-Small | 0.697 , 0.073 | 0.823 , 0.031 | 0.901 , 0.032 | 0.949 , 0.010 |
| EfficientNet-B3 | 0.624 , 0.030 | 0.778 , 0.014 | 0.841 , 0.035 | 0.928 , 0.007 |
| DenseNet-121 | 0.614 , 0.091 | 0.769 , 0.046 | 0.821 , 0.055 | 0.924 , 0.025 |

architectural inductive biases influence DR severity grading performance.

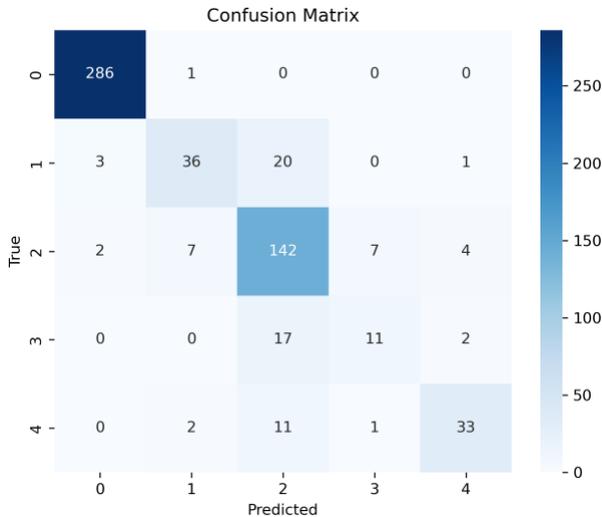

Figure 2: Confusion matrix for the best-performing single backbone (ResNet-50) on five-class DR grading.

Table 1 summarizes the overall performance, and Figure 2 highlights the dominant error modes, showing that most misclassifications occur between neighboring grades. Overall, the results indicate that modern CNN backbones provide the strongest single-model baselines, with ResNet-50 and ConvNeXt-Tiny emerging as the best-performing pairs. This pattern is consistent with the advantage of convolutional inductive bias for capturing fine-grained and localized retinal features, such as microaneurysms, hemorrhages, and small exudative lesions, which are central to DR grading. ConvNeXt, in particular, appears to benefit from modern training strategies while preserving the locality of convolutional representations, making it especially suitable when lesion evidence is subtle and spatially sparse.

Transformer-based backbones, including the Swin Transformer and DeiT-Small, exhibit competitive ordinal behavior, suggesting that attention mechanisms are effective at capturing global retinal context and disease severity progression. However, their class-balanced performance is more affected by the difficulty in the minority grades, as reflected by the gap between ordinal agreement and macro-level classification results. This is not unexpected in DR grading, where distinguishing later disease stages often depends on both the broad vascular context and reliable identification of sparse but clinically important findings. In such settings, transformers may model the global structure well; however, they still face challenges when evidence for minority grades is limited, noisy, or only partially visible.

In contrast, EfficientNet-B3 and DenseNet-121 are weaker baselines under the same experimental protocol, suggesting lower robustness under the combined effects of heterogeneous image quality and pronounced class imbalance. Although both architectures are well established in medical imaging, their greater sensitivity to training dynamics and imbalance may contribute to stronger degradation in minority-grade classification when evaluated under identical conditions.

Across all backbones, the main failure pattern is consistent with the overall study objective: misclassifications occur primarily between adjacent grades rather than between distant stages, whereas minority and late-stage classes remain the main sources of difficulty. Importantly, this backbone-level comparison also guides the subsequent analysis in RQ2 by identifying a strong and complementary set of models for ensemble construction. Based on these results, ResNet-50 and ConvNeXt-Tiny are forwarded as the most reliable single-model components for ensembling, whereas Swin and DeiT provide a transformer perspective that helps interpret whether gains arise from local lesion sensitivity or global context modeling.

> ✅ **Answer to RQ1**
> ResNet-50 and ConvNeXt-Tiny provided the strongest single-model baselines for five-class DR grading under the unified protocol, reaching cross-validated QWK values up to 0.919 and 0.914, respectively. This indicates that, in the present setting, modern CNN-based architectures offered the most reliable standalone performance.

### RQ2: Ensemble Performance

RQ2 examines whether combining complementary backbones improves robustness beyond single-model baselines and whether class-wise fusion provides additional benefits for ordinal DR grading. Guided by the backbone comparison in RQ1, we built ensembles from the two best-performing single models, ResNet-50 and ConvNeXt-Tiny, and compared four fusion strategies:

hard voting, weighted soft voting, stacking with a logistic regression meta-learner, and a proposed hybrid class-level fusion scheme.

Across the five folds, probability-level fusion methods are more stable than discrete voting. As summarized in Table 2, weighted soft voting achieves the best average ordinal agreement and the lowest fold-to-fold variability, indicating that global probability fusion provides reliable gains under class imbalance. Stacking yields comparable performance but does not exceed weighted soft voting on average, suggesting that the learned meta-combination offers limited additional benefits in this setting.

Table 2: QWK summary across five folds (mean , standard deviation). Higher is better.

| Method | QWK (mean , SD) |
|---|---|
| Hard voting | 0.9255 , 0.0226 |
| Weighted soft voting | **0.9337 , 0.0172** |
| Stacking | 0.9256 , 0.0211 |
| Hybrid (class-level) | 0.9282 , 0.0242 |

The hybrid class-level ensemble achieves a competitive mean QWK but exhibits higher sensitivity to fold composition. Table 3 shows that hybrid fusion matches or exceeds alternative strategies in some folds but underperforms in one fold, where the QWK decreases notably, increasing the overall variance. This behavior is consistent with class-wise specialization: class-specific competence estimates are derived from validation behavior and can be less stable for minority grades when the validation sample size is limited.

Table 3: Fold-wise QWK for ensemble methods under five-fold cross-validation.

| Fold | Hard voting | Soft voting | Stacking | Hybrid |
|---|---|---|---|---|
| 0 | 0.908 | 0.926 | 0.915 | 0.924 |
| 1 | 0.907 | 0.919 | 0.911 | 0.935 |
| 2 | 0.924 | 0.935 | 0.931 | 0.924 |
| 3 | 0.925 | 0.926 | 0.910 | 0.896 |
| 4 | 0.963 | 0.963 | 0.960 | 0.963 |

To test whether hybrid fusion yields a consistent improvement over standard ensembles, we performed paired Wilcoxon signed-rank tests on fold-wise QWK with the Holm–Bonferroni correction. As reported in Table 4, none of the comparisons (hybrid vs. hard voting, soft voting, stacking) are statistically significant after correction, and the bootstrap confidence intervals for mean QWK differences include zero. Overall, these results indicate that ensembling improves ordinal agreement, with weighted soft voting providing the most consistent gains in this cross-validation setting, whereas hybrid class-level fusion remains a plausible specialization approach without a reliable advantage over global probability fusion.

*Grade-Wise Performance Patterns:* To characterize grade-wise robustness beyond aggregate metrics, we report class-wise precision, recall, and F1-score for the weighted soft voting ensemble under five-fold cross-validation (Table 5). The performance is strongest for Class 0 (no DR), reflecting the clear separation between normal and pathological images in this dataset. Among the pathological grades, Class 2 (moderate) appears to be the most stable, with a relatively high recall and low fold-to-fold variability. This suggests that moderate-stage retinal features are recognized more consistently across data splits. In contrast, the poorest class-wise performance is observed for Class 3 (severe), where both recall and F1 show substantial variations across folds. This behavior is consistent with the known challenge of separating severe NPDR from neighboring grades under conditions of class imbalance and variable image quality. Classes 1 and 4 fall in between, but their higher variability relative to Class 2 indicates that mild disease signs and proliferative features remain less consistently captured across splits.

Overall, ensembling improves ordinal agreement compared with single backbones. Among the evaluated strategies, weighted soft voting provides the most consistent gains in QWK across folds, whereas hybrid class-level fusion is competitive but more sensitive to fold composition and does not show a statistically reliable advantage over standard fusion methods under the current cross-validation setting.

> ✅ **Answer to RQ2**
> Ensembling improved ordinal agreement beyond individual backbones, with weighted soft voting yielding the most consistent performance across folds (QWK = 0.934 , 0.017). Although the hybrid class-level ensemble was competitive, paired fold-wise comparisons did not show a statistically reliable improvement over standard fusion strategies.

### RQ3: Multimodal Explainability Evaluation

*Visual and Textual Explanation Analysis:* To qualitatively address RQ3, we examined whether the explainability pipeline provides (i) visually plausible evidence localization and (ii) concise textual rationales that are consistent with the predicted DR grade. To maintain focus on the presentation, we report one representative tripanel per *unique* severity class available in our qualitative set. Each tripanel combines the original fundus photograph, the Grad-CAM/Grad-CAM++ overlay for the predicted class (with the predicted probability vector), and the VLM-generated explanation.

Figure 3 illustrates a case with no DR and near-complete confidence in class 0. The generated explanation highlights the absence of microaneurysms, hemorrhages, and exudates, which is consistent with the predicted grade. The Grad-CAM++ overlay appears relatively diffuse and partly attends to prominent anatomical regions, such as the optic disc, suggesting that the prediction is influenced more by the overall normal retinal appearance than by localized lesion-specific cues.

Table 4: Paired statistical comparison of QWK (Hybrid vs baselines).

| Comparison | Mean ΔQWK | 95% CI | Wilcoxon p / Holm p |
|---|---|---|---|
| Hybrid vs Hard vote | +0.003 | [*0.015, 0.017] | 1.000 / 1.000 |
| Hybrid vs Soft vote | *0.006 | [*0.019, 0.007] | 0.625 / 1.000 |
| Hybrid vs Stacking | +0.003 | [*0.009, 0.014] | 0.813 / 1.000 |

Table 5: Class-wise performance of the weighted soft voting ensemble under five-fold cross-validation (mean , SD).

| Class | Precision | Recall | F1-score |
|---|---|---|---|
| 0 (No DR) | 0.983 , 0.012 | 0.997 , 0.008 | 0.990 , 0.007 |
| 1 (Mild) | 0.834 , 0.104 | 0.683 , 0.149 | 0.740 , 0.085 |
| 2 (Moderate) | 0.783 , 0.033 | 0.883 , 0.044 | 0.829 , 0.023 |
| 3 (Severe) | 0.720 , 0.228 | 0.467 , 0.217 | 0.530 , 0.203 |
| 4 (Proliferative) | 0.832 , 0.159 | 0.696 , 0.213 | 0.739 , 0.133 |

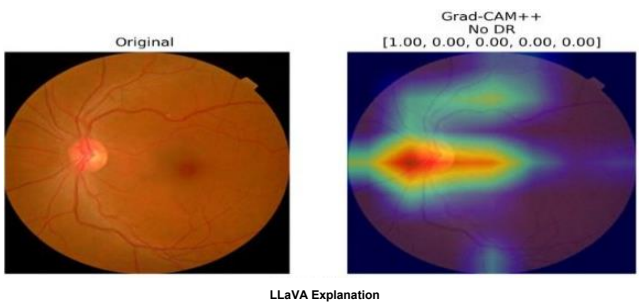

**User Input:** Fundus photograph. The classifier predicted No DR diabetic retinopathy (class 0) with probabilities [1.00, 0.00, 0.00, 0.00, 0.00]. In 2–4 sentences, describe the key retinal findings that support this grade, using clinical terms (e.g., microaneurysms, dot-blot hemorrhages, exudates, neovascularization, macular involvement, vessel changes). Be specific and avoid speculation.

**Assistant:** The key retinal findings that support the grade of No DR diabetic retinopathy include the absence of microaneurysms, dot-blot hemorrhages, exudates, neovascularization, macular involvement, and vessel changes. These findings suggest that the retina appears to be normal, without any signs of diabetic retinopathy.

Figure 3: Tripanel example for Class 0 (No DR): original image, Grad-CAM++ overlay with predicted probabilities, and VLM explanation.

Figure 4 presents a representative example predicted as Moderate DR. The attribution overlay is concentrated over regions that plausibly correspond to lesion-related evidence, while the generated explanation uses terminology consistent with this grade, including microaneurysms and hemorrhages. This case illustrates the intended behavior of the proposed pipeline: coarse visual evidence localization combined with a brief clinician-oriented rationale that aligns with the predicted grade.

Figure 5 presents a Severe DR case. Relative to lower grades, the attribution map appears stronger and more localized, consistent with the presence of more pronounced pathological changes. The generated explanation also conveys a greater disease burden by

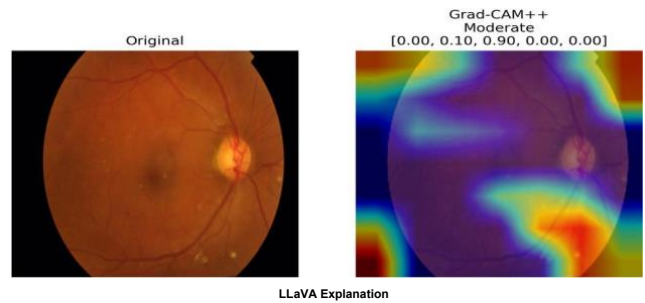

**User Input:** This is a retinal fundus image. The AI predicted Moderate diabetic retinopathy (class 2) with probabilities [0.00, 0.10, 0.90, 0.00, 0.00]. Describe the retinal signs (microaneurysms, hemorrhages, exudates, vessels, macula) that support this diagnosis in concise clinical terms.

**Assistant:** The image depicts a close-up of a person's eye, showing a retinal fundus view. The AI has predicted moderate diabetic retinopathy, which is characterized by the presence of microaneurysms and hemorrhages in the retina.

Figure 4: Tripanel example for Class 2 (Moderate DR): original image, Grad-CAM++ overlay with predicted probabilities, and VLM explanation.

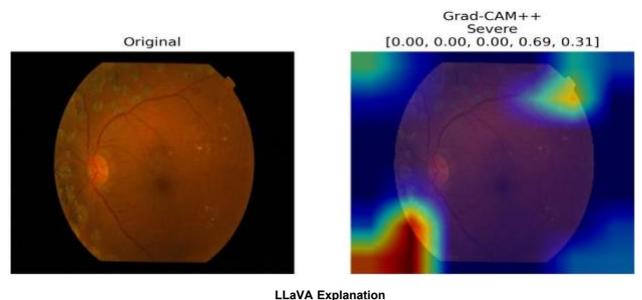

**User Input:** This is a retinal fundus image. The AI predicted Severe diabetic retinopathy (class 3) with probabilities [0.00, 0.00, 0.00, 0.69, 0.31]. Describe the retinal signs (microaneurysms, hemorrhages, exudates, vessels, macula) that support this diagnosis in concise clinical terms.

**Assistant:** In the image, there are multiple signs of diabetic retinopathy. The retina is covered in microaneurysms, which are small blood vessel leaks, and there are hemorrhages present as well. Exudates are also visible, which are fluid accumulations in the retina.

Figure 5: Tripanel example for Class 3 (Severe DR): original image, Grad-CAM++ overlay with predicted probabilities, and VLM explanation.

using terminology appropriate for severe-stage findings. Overall, this qualitative pattern suggests that when pathological cues are more clearly visible, both visual localization and the textual rationale become more specific and decisive.

Figure 6 shows a Proliferative DR case. The generated explanation includes late-stage descriptors, including neovascularization and advanced hemorrhagic changes, that align with the predicted grade. This example demonstrates the benefit of multimodal explainability: the heatmap provides a coarse view of the regions contributing to the decision, whereas the text conveys a concise clinical interpretation of the predicted severity.

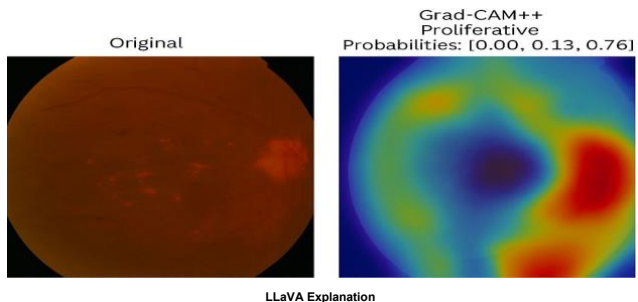

**LLaVA Explanation**

**User Input:** This is a retinal fundus image. The AI predicted Proliferative DR diabetic retinopathy (class 4) with probabilities [0.00, 0.12, 0.13, 0.00, 0.76]. Describe the retinal signs (microaneurysms, hemorrhages, exudates, vessels, macula) that support this diagnosis in concise clinical terms.

**Assistant:** The retinal fundus shows advanced diabetic changes consistent with proliferative diabetic retinopathy. Multiple microaneurysms and dot-blot hemorrhages are scattered across the posterior pole. There is evidence of neovascularization near the optic disc and along the vascular arcades, suggesting new fragile vessel formation. Hard exudates are visible in the macular region, indicating lipid leakage. These findings collectively suggest a severe, vision-threatening stage of diabetic retinopathy that requires urgent clinical intervention.

Figure 6: Tripanel example for Class 4 (Proliferative DR): original image, Grad-CAM++ overlay with predicted probabilities, and VLM explanation.

Overall, the class-wise examples demonstrate that the pipeline can produce grade-consistent explanations by pairing coarse evidence localization with short clinical rationales. The remaining limitations are most apparent in lower grades, in which attribution can be diffuse and explanations may rely on the absence of findings, motivating the quantitative evaluation of semantic fidelity, image–text alignment, and clinical term coverage in the next subsection.

*Quantitative Evaluation of Explanations:* To complement the qualitative inspection, we quantitatively evaluate the generated textual rationales using three complementary criteria: semantic fidelity to grade-specific reference templates, image–text alignment, and clinical term coverage. The results for both VLM variants are presented in Table 6. For completeness, we also report the underlying grading performance associated with the same evaluation set.

Overall, both VLM variants produce explanations with comparable image–text alignment, as reflected by similar CLIPScore values. The biomedical-adapted model achieves higher coverage, indicating that its explanations include a larger fraction of the predefined clinical descriptor vocabulary. In contrast, the lightweight LLaVA model attains a higher BERTScore against grade-specific reference templates, suggesting stronger semantic similarity to the canonical template phrasing. Taken together, these findings suggest a trade-off between *clinical completeness* (coverage) and *template-level semantic similarity* (BERTScore), whereas visual grounding remains broadly comparable between the two models. Importantly, the classifica-tion metrics, including accuracy, macro-F1, kappa, and AUC, remained unchanged across both settings, confirming that the observed differences arise from the explanation generator rather than from any change in the grading pipeline.

In practical terms, the higher coverage obtained by llava-med-v1.5-mistral-7b suggests that this model more consistently includes clinically relevant concepts, which may be advantageous for clinician-facing summaries. By contrast, the higher BERTScore of llava-hf/llava-1.5-1b-hf indicates closer similarity to the grade-specific reference templates, possibly reflecting more Because the CLIPScore is similar across both models, the main differences seem to concern language content and terminology rather than image–text compatibility. Together, these findings motivate the reporting of both fidelity-oriented and completeness-oriented metrics when evaluating explanation generators.

Overall, the qualitative examples show that Grad-CAM and Grad-CAM++ provide plausible visual evidence of localization across grades; however, they remain coarse attribution signals that may overlap with anatomical structures or artifacts and should not be interpreted as lesion segmentations. The VLM-generated rationales are generally consistent with the predicted grade and use clinically meaningful terminology, with conservative prompting helping to reduce overconfident or weakly grounded statements. Quantitatively, llava-med-v1.5-mistral-7b achieves higher clinical term coverage, whereas llava-hf/llava-1.5-1b-hf yields higher template-based semantic similarity (BERTScore), indicating a trade-off between clinical completeness and template-level phrasing. Importantly, the classification performance remains unchanged across VLM variants, confirming that the differences are attributable to the explanation generator rather than the grading model.

> ✅ **Answer to RQ3**
> The proposed multimodal explanation framework produced plausible Grad-CAM++ localizations and grade-consistent VLM rationales. Nevertheless, the evaluation showed a trade-off between clinical coverage and template-level semantic similarity, with comparable image–text alignment but clear room for improvement in explanation grounding and reliability.

Table 6: Quantitative comparison of textual explanation quality for the two VLM variants. Classification metrics are reported for the same evaluation set.

| VLM | Acc. | Macro-F1 | Kappa | AUC | BERTScore | Coverage | CLIPScore |
|---|---|---|---|---|---|---|---|
| llava-med-v1.5-mistral-7b | 0.844 | 0.685 | 0.752 | 0.950 | 0.048 | **0.700** | 0.344 |
| llava-hf/llava-1.5-1b-hf | 0.844 | 0.685 | 0.752 | 0.950 | **0.072** | 0.602 | 0.340 |

## 5. Discussion

The results show a consistent pattern across backbone comparison, ensembling, and multimodal explanation analysis. Modern CNN backbones, particularly ResNet-50 and ConvNeXt-Tiny, provided the strongest overall balance between class-balanced performance and ordinal agreement, while transformer-based models remained competitive but somewhat less consistent on minority grades. Ensembling further improved robustness, with weighted soft voting producing the most stable gains across folds, whereas the hybrid class-level fusion strategy was competitive but did not demonstrate a statistically significant advantage in paired testing. At the explanation level, Grad-CAM++ provided plausible but coarse evidence localization, and VLM-generated rationales were generally consistent with the predicted grade under constrained prompting. Quantitatively, the evaluated VLM variants revealed a trade-off between clinical term coverage and semantic similarity to reference templates, while showing similar CLIPScore values and leaving classification performance unchanged.

The backbone comparison suggests that convolutional inductive bias remains highly effective for DR severity grading, where the most informative cues often appear as small, localized lesions distributed across the retina. ResNet-50 and ConvNeXt-Tiny likely benefit from strong local feature extraction together with stable optimization under transfer learning and augmentation. ConvNeXt, in particular, combines modern training practices with preserved convolutional locality, which may help maintain performance when lesion evidence is sparse or partly obscured by acquisition artifacts.

Transformer-based models provide a complementary perspective. Swin and DeiT are well suited to capturing global context and long-range dependencies, which can be important when disease severity is reflected in broader vascular changes, diffuse hemorrhages, or more global structural patterns. However, the gap between their ordinal agreement and macro-averaged performance is consistent with the challenge of distinguishing minority grades. Even when the overall severity ordering is learned reasonably well, separating adjacent stages remains difficult when high-grade findings are subtle, rare, or inconsistently represented in a single image. This is also consistent with the broader pattern observed across models, where most errors occurred between neighboring grades rather than as extreme grade jumps. From a clinical standpoint, this is expected, as the boundaries between mild, moderate, and severe DR depend on lesion counts and spatial distribution, which are difficult to infer reliably from heterogeneous images without explicit lesion counting or multi-view information.

> **Model Selection Insight..** For dependable DR grading under standard screening conditions, strong CNN baselines such as ResNet-50 and ConvNeXt-Tiny remain the most reliable choices in the present study, offering the best balance between ordinal agreement and class-sensitive performance.

Ensembling is often motivated by the idea that different models make different mistakes, especially near class boundaries. In this study, global probability fusion through weighted soft voting produced the most consistent improvement in ordinal agreement across folds. This is a plausible outcome for an imbalanced ordinal task, as probability averaging can reduce variance, smooth unstable predictions near grade boundaries, and limit overconfident errors from any single backbone. The similar performance of stacking suggests that a learned meta-combiner does not necessarily offer additional benefits beyond well-calibrated probability fusion, particularly when the base learners are already strong and the amount of fold-specific validation data is limited.

The hybrid class-level ensemble was designed to exploit grade-wise complementarity by allowing the preferred model to vary across classes. The fold-wise results suggest that this approach can match or, in some splits, exceed standard fusion; however, it can also underperform when class-wise competence estimates are unstable. This is especially relevant for minority grades, where small changes in fold composition can substantially affect the reliability of class-specific statistics. The paired Wilcoxon tests with the Holm correction did not show statistically significant differences between hybrid fusion and the baseline methods, and the confidence intervals for the mean differences included zero. Under the current cross-validation setting, this indicates that hybrid class-level fusion does not provide a consistent advantage over global probability fusion.

> **Ensemble Selection Insight..** For dependable DR grading under standard screening conditions, strong CNN baselines such as ResNet-50 and ConvNeXt-Tiny remain the most reliable choices in the present study, offering the best balance between ordinal agreement and class-sensitive performance.

A quantitative evaluation of explanations is essential if multimodal interpretability is to move beyond illustrative examples; however, such evaluations must be interpreted carefully in medical applications. In this study, we used complementary metrics that capture different aspects of explanation quality: semantic similarity to grade-specific templates through BERTScore, image–text alignment through CLIPScore, and keyword-based clinical term coverage. The results demonstrate that the two VLM variants behave differently in predictable ways. The biomedical-adapted model achieved higher coverage, suggesting that it more consistently included clinically relevant descriptors from the predefined vocabulary. The lightweight model achieved a higher BERTScore against the grade templates, indicating closer alignment with canonical reference phrasing. The CLIPScore was similar for both models, implying that image–text compatibility was broadly comparable, and that the main differences were linguistic rather than visual.

These findings highlight two important caveats. First, template-based semantic similarity is sensitive to phrasing and does not necessarily reflect clinical correctness. A model may achieve a higher BERTScore simply by producing language that more closely resembles the template, without necessarily providing richer or more image-specific reasoning. Second, coverage serves only as a proxy for completeness and not as a guarantee of faithfulness. The inclusion of clinical terms does not ensure that the referenced findings are truly present or correctly localized. For this reason, the three metrics should be interpreted together: semantic similarity provides a reference-based signal, CLIPScore offers a reference-free grounding signal, and coverage reflects whether the explanation includes clinically meaningful concepts.

The fact that the classification metrics remained identical across the two VLM settings is also important. This confirms that the grading model itself was unchanged and that the evaluation isolates differences at the explanation stage. This supports the methodological claim that explanation generation can be treated as a separate and measurable output layer, one that can be improved independently of classification performance.

> **Clinical Use Insight.** The proposed multimodal explanation pipeline is best interpreted as a decision-support layer rather than as a standalone source of clinical evidence: Grad-CAM(++) can indicate where the model is focusing, and VLM-generated rationales can improve readability, but both still require expert review before practical use.

### Clinical and Practical Implications

From a clinical and screening perspective, these findings suggest that the proposed framework is most useful as a decision-support tool rather than as a replacement for expert judgment. In a practical workflow, the models predicted DR grade could support case prioritization by helping graders or ophthalmologists identify images that may require faster review, referral, or closer follow-up. For example, when the system predicts severe or proliferative DR with high confidence, this may help flag a case for urgent specialist attention. Conversely, in images predicted as no DR or mild DR, the output may support routine review, while still allowing the clinician to verify that subtle lesions have not been overlooked. In this sense, the main value of the system is not only in producing a grade, but also in providing additional visual and textual cues that can help experts assess whether the prediction appears clinically plausible.

> **Screening Workflow Insight.** In screening-oriented use, the framework is best viewed as a triage and review aid: the predicted grade can help prioritize cases, while the accompanying explanations can support a faster and more transparent expert assessment.

The explainability component is especially relevant in this regard. Grad-CAM(++) visualizations can help indicate whether the model is focusing on lesion-rich retinal regions rather than on irrelevant structures or image artefacts. For instance, if a moderate DR prediction is accompanied by activation over areas containing hemorrhages or exudate-like regions, the explanation may increase the clinicians confidence that the model is attending to meaningful evidence. Similarly, VLM-generated rationales can translate the model output into short clinician-readable descriptions, which may be useful during image review or documentation-oriented workflows. For example, a rationale stating that the prediction is associated with multiple hemorrhagic or exudative findings may be easier to inspect and interpret than a heatmap alone, particularly for non-technical graders who are less familiar with saliency maps. In this way, the combination of grade prediction, visual localization, and textual rationale may help bridge the gap between algorithmic output and expert review.

☿ **Expert-in-the-Loop Role.** The combined use of grade prediction, visual localization, and short textual rationale may be particularly helpful during review, because it gives clinicians and graders more than a single label and makes the model output easier to inspect in context.

At the same time, our results also clarify the limits of such assistance. Because Grad-CAM(++) remains coarse, it should not be interpreted as lesion-level confirmation, and because VLM-generated text is evaluated only indirectly through similarity, coverage, and image–text alignment metrics, it cannot be assumed to be fully faithful in every case. This is particularly important in borderline images, in minority severity grades, or in cases where disease staging depends on subtle lesion counts and spatial distribution. In such settings, a clinician should treat the explanation as a prompt for closer inspection rather than as direct evidence. For example, if the model predicts severe DR but the highlighted regions are diffuse or the generated rationale is generic, this would signal the need for careful manual reassessment rather than immediate trust in the automated output.

More broadly, the results suggest that the framework could be useful in settings where screening volume is high and interpretability matters. In a teleophthalmology or population-screening context, the model could assist first-pass triage, while the accompanying explanations help human reviewers decide whether the case appears straightforward or whether the prediction should be questioned. For expert users, the system may serve as an additional second reader that highlights potentially relevant regions and summarizes possible retinal findings. For less specialized graders, the textual explanation may improve transparency by making the model decision easier to interpret. However, the system is not yet suited to autonomous clinical reporting. Its most realistic role, based on the present results, is to support expert-centered review by combining competitive grading performance with explanation outputs that can be inspected, questioned, and validated by clinicians.

🏛 **Expert-in-the-Loop Role.** The framework should not be treated as an autonomous reporting system. Its most realistic near-term role is as an expert-in-the-loop support tool, especially in high-volume screening settings where prioritization and interpretability are both important.

### Comparison with Prior Work

To place the reported performance in context, we compare our results with representative recent studies on DR severity grading and explainability, as summarized in Table 7. However, direct numerical comparison should be interpreted cautiously, as prior work often relies on different train/test splits, DR-specific external pretraining, additional preprocessing steps, or different evaluation sets, whereas our results are reported as the mean , SD under stratified five-fold cross-validation. With this caveat, our best-performing ensemble (weighted soft voting) achieved strong ordinal agreement (QWK mean ù 0.934) and high macro-AUC, which is broadly consistent with the performance range reported by recent DR grading pipelines that emphasize transfer learning and ensemble strategies on APTOS-like benchmarks. For example, Dharrao et al. report a QWK of approximately 0.923 with a fixed train/test setting for an EfficientNet-based pipeline and provide Grad-CAM visualizations for qualitative interpretability support (Dharrao et al., 2025). Akhtar et al. report a QWK of approximately 0.93 for stage grading in their dual-branch transfer learning approach (Akhtar et al., 2025). Chilukoti et al. report a higher QWK on APTOS by combining DR-specific pretraining with an ensemble strategy (Chilukoti et al., 2024), suggesting that cross-dataset pretraining and task-tailored training procedures can further improve ordinal agreement beyond standard ImageNet initialization.

From an interpretability standpoint, much of the DR grading literature relies mainly on post hoc heatmaps such as CAM and Grad-CAM as qualitative evidence. Hybrid CNN–Transformer approaches often include attribution maps to improve transparency but do not systematically quantify explanation qual-ity (Xu et al., 2024). In contrast, our study treats explanation generation as a measurable output by combining Grad-CAM(++) localization with VLM-generated rationales and by reporting complementary quantitative measures that capture semantic fidelity to grade templates, image–text alignment, and clinical term coverage. This complements emerging retinal vision–language directions that aim to couple predictive models with language-based accountability; however, these approaches are typically evaluated on different datasets or tasks (La Rosa et al., 2026; This et al., 2025).

Taken together, the comparison with prior work reinforces the practical picture emerging from our results. Strong CNN baselines and weighted soft-voting ensembles provide a reliable foundation for DR grading under the present setting, whereas the explainability pipeline adds value primarily as an interpretable support layer rather than as an autonomous clinical reporting mechanism.

### Limitations

This study has several limitations that should be considered when interpreting the results. First, the evaluation was conducted on a single public dataset with heterogeneous image quality and a fixed label taxonomy. Although cross-validation reduces sensitivity to a single split, it does not replace external validation on independent cohorts or on images acquired with different devices and protocols. Second, the most clinically challenging grades are also the least represented, which limits both classifier learning and the stability of class-wise competence estimation for hybrid fusion. Third, the explanation metrics rely on reference templates and automated scoring. APTOS does not provide ground-truth rationales, and template-based evaluation cannot fully capture clinical correctness, specificity, or faithfulness to visible evidence. Fourth, Grad-CAM provides only coarse attribution maps and is not intended for lesion-level localization. In the absence of segmentation ground truth, the fidelity of these localization cues cannot be rigorously assessed. Finally, although constrained prompting can reduce the risk of hallucination, VLM outputs remain probabilistic and may still be sensitive to prompt formulation and decoding settings. Therefore, expert review remains necessary before any clinical deployment.

Table 7: Comparison with representative studies on DR grading and explainability. QWK and evaluation protocols are reported as stated in the cited works; differences in datasets and splits can limit direct comparability. NR = not reported.

| Study | Dataset(s) | Task | Protocol | Model / Fusion | QWK | Explainability / Text | Key Differences |
|---|---|---|---|---|---|---|---|
| **This work** | APTOS 2019 | 5-class grading | Stratified 5-fold CV (mean,SD) | Backbone benchmarking + weighted soft voting (best) | 0.934, 0.017 | Grad-CAM(++) + VLM rationales; quantitative text evaluation | Joint evaluation of grading and multimodal explainability; includes quantitative assessment of generated text using BERTScore, Coverage, and CLIPScore. |
| Chilukoti et al. (Chilukoti et al., 2024) | EyePACS, APTOS, Messidor-2 | 5-class grading | Transfer learning across datasets (as reported) | EfficientNet-based ensemble; DR-specific pretraining | 0.967 (APTOS) | Focus on grading; no VLM text evaluation reported | Stronger reported QWK on APTOS, but direct comparison is limited by different data protocol and cross-dataset transfer setting. Does not evaluate multimodal textual explanations. |
| Dharrao et al. (Dharrao et al., 2025) | APTOS 2019 | 5-class grading | Cross-validation + held-out test (as reported) | EfficientNetB0 with attention gates | 0.923, 0.008 | Grad-CAM-style visual interpretability; no text-rationale evaluation | Uses attention-enhanced single-model design. Our study adds systematic backbone benchmarking, ensemble comparison, and quantitative text-based explanation analysis. |
| Mohsen et al. (Mohsen et al., 2025) | APTOS 2019, DDR | 5-class grading | 80/20 train/test; 10% of train for validation (as reported) | RadFuse (fundus + RadEx representation fusion) | 0.932 | Model comparison and error analysis; no VLM text evaluation reported | Performance is close to ours, but the method focuses on representation fusion rather than multimodal explainability. Evaluation protocol differs from stratified 5-fold CV. |
| Sun et al. (Sun et al., 2025) | DDR, Messidor-2, APTOS 2019 | 5-class grading | Per-class 6:2:2 train/val/test split (as reported) | MAFNet (multi-scale + attention modules) | 0.936 (APTOS) | No VLM text; focuses on architectural attention modules | Architectural emphasis is on multi-scale feature extraction and attention. Our work contributes comparative benchmarking plus explanation generation and explicit explanation-quality metrics. |
| Xu et al. (Xu et al., 2024) | APTOS 2019 | DR classification (task framing differs) | Protocol differs (as reported in paper) | Hybrid EfficientNet + Swin Transformer | NR | Class activation maps (CAM) for qualitative visualization | Relevant for hybrid CNN-transformer design, but lacks reported QWK and does not assess textual explanations or explanation quality quantitatively. |
| Akhtar et al. (Akhtar et al., 2025) | Messidor-1 | Stage grading (4-stage / 2-stage) | Protocol differs (as reported in paper) | RSG-Net (custom CNN) | 0.930 (stage grading) | Reports localization/abnormality cues; no VLM text evaluation | Not directly comparable because of different dataset and grading formulation. Our work addresses 5-class APTOS grading and multimodal interpretability under a unified protocol. |

**Future Directions**

Several directions may help strengthen the clinical reliability and broader applicability of the proposed pipeline. First, the proposed pipeline should be validated on additional DR datasets and real-world screening data, especially under conditions of device shift and variable image quality. Second, class-wise ensembling could be made more stable through repeated cross-validation, regularized weight estimation, competence measures based on calibrated probabilities, or ordinal-aware fusion objectives that explicitly penalize large grading errors. Third, explanation grounding could be strengthened by incorporating lesion segmentation or detection modules or by linking attribution maps to explicit lesion concept extraction to reduce ambiguity. From an evaluation perspective, ophthalmologist-in-the-loop assessment remains important for judging both grading reliability and explanation faithfulness and could be complemented by structured annotation of key findings rather than relying solely on template-based references. Finally, practical deployment would require uncertainty monitoring and reporting; future work should therefore integrate calibrated uncertainty estimates and guardrails to reduce overconfident explanations when the underlying evidence is limited or visually unclear.

## 6. Conclusion

This study presents a unified pipeline for five-class diabetic retinopathy grading that combines strong image-based classifiers with multimodal explainability. Under a controlled evaluation protocol on the APTOS 2019 dataset, modern CNN backbones, particularly ResNet-50 and ConvNeXt–Tiny, emerged as the most reliable single-model baselines, whereas transformer backbones remained competitive but were somewhat less consistent under class imbalance. Ensembling further improved ordinal agreement, with weighted soft voting producing the most stable gains across folds. Although hybrid class-level fusion was also competitive, it did not show a statistically consistent advantage within the present cross-validation setting. Beyond grading performance, this study also treated explanation generation as a measurable output rather than a purely illustrative add-on. Grad-CAM and Grad-CAM++ provided visually plausible, though coarse, evidence localization, whereas constrained prompting enabled VLMs to generate textual rationales that were generally consistent with the predicted grade. A quantitative comparison of the two VLM variants revealed a practical trade-off between clinical completeness, reflected in term coverage, and template-level semantic similarity, reflected in BERTScore, while image-text alignment remained similar across models. Overall, these findings support the use of multimodal explainability as a complementary layer for improving transparency in DR grading. At the same time, they highlight the need for external validation and expert assessment before clinical faithfulness and deployment readiness can be established.

## Data and code availability

The APTOS 2019 Blindness Detection dataset Karthik et al. (2019) is publicly available on Kaggle under its original terms of use. The code used in this study will be released publicly at:
https://github.com/PirBakhsh0708/dr_grading_vlm.

## Acknowledgments

This work has been partially supported by the European Union through the Italian Ministry of University and Research, Project PNRR *"D3-4Health: Digital Driven Diagnostics, Prognostics and Therapeutics for Sustainable Health Care"*, PNC 0000001; CUP B53C22006090001.

## References

Aburass, S., Dorgham, O., Al Shaqsi, J., Abu Rumman, M., Al-Kadi, O., 2025. Vision transformers in medical imaging: a comprehensive review of advancements and applications across multiple diseases. Journal of Imaging Informatics in Medicine 38, 3928–3971.

Akhtar, S., Aftab, S., Ali, O., Ahmad, M., Khan, M.A., Abbas, S., Ghazal, T.M., 2025. A deep learning based model for diabetic retinopathy grading. Scientific reports 15, 3763.

Boulaabi, M., Gader, T.B.A., Echi, A.K., Bouraoui, Z., 2025. Enhancing dr classification with swin transformer and shifted window attention, in: International Conference on Artificial Intelligence in Medicine, Springer. pp. 57–61.

Buslaev, A., Iglovikov, V.I., Khvedchenya, E., Parinov, A., Druzhinin, M., Kalinin, A.A., 2020. Albumentations: fast and flexible image augmentations. Information 11, 125.

Interpretable DR Grading with Vision-Language ModelsCheng, Z., Ong, A.Y., Wagner, S.K., Merle, D.A., Ju, L., Zhang, H., Chen, R., Pang, L., Li, B., He, T., et al., 2025. Understanding the robustness of vision-language models to medical image artefacts. NPJ Digital Medicine 8, 727.

Chilukoti, S.V., Shan, L., Tida, V.S., Maida, A.S., Hei, X., 2024. A reliable diabetic retinopathy grading via transfer learning and ensemble learning with quadratic weighted kappa metric. BMC Medical Informatics and Decision Making 24, 37.

Chong, D.D., Das, N., Singh, R.P., 2024. Diabetic retinopathy: Screening, prevention, and treatment. Cleveland Clinic journal of medicine 91, 503–510.

Damm, H., Pakull, T.M., Becker, H., Bracke, B., Eryilmaz, B., Bloch, L., Brüngel, R., Schmidt, C.S., Rückert, J., Pelka, O., et al., 2025. Overview of imageclefmedical 2025–medical concept detection and interpretable caption generation. Pro-ceedings of the CLEF, Madrid, Spain , 9–12.

Dharrao, D., Dharrao, M., Patil, S., Salvin, S., Ahire, P., Dongre, Y., 2025. Ai-driven detection and classification of diabetic retinopathy stages using efficientnetb0. Discover Applied Sciences .

Goh, J.H.L., Ang, E., Srinivasan, S., Lei, X., Loh, J., Quek, T.C., Xue, C., Xu, X., Liu, Y., Cheng, C.Y., et al., 2024. Comparative analysis of vision transformers and conventional convolutional neural networks in detecting referable diabetic retinopathy. Ophthalmology Science 4, 100552.

He, K., Zhang, X., Ren, S., Sun, J., 2016. Deep residual learning for image recognition, in: Proceedings of the IEEE conference on computer vision and pattern recognition, pp. 770–778.

Hessel, J., Holtzman, A., Forbes, M., Le Bras, R., Choi, Y., 2021. Clipscore: A reference-free evaluation metric for image captioning, in: Proceedings of the 2021 conference on empirical methods in natural language processing, pp. 7514–7528.

Holland, R., Taylor, T.R., Holmes, C., Riedl, S., Mai, J., Patsiamanidi, M., Mitsopoulou, D., Hager, P., Müller, P., Paetzold, J.C., et al., 2025. Specialized curricula for training vision language models in retinal image analysis. NPJ Digital Medicine 8, 532.

Holm, S., 1979. A simple sequentially rejective multiple test procedure. Scandinavian journal of statistics , 65–70.

Huang, G., Liu, Z., Van Der Maaten, L., Weinberger, K.Q., 2017. Densely connected convolutional networks, in: Proceedings of the IEEE conference on computer vision and pattern recognition, pp. 4700–4708.

Huang, X., Ai, Z., She, C., Li, Q., Wei, Q., Xu, S., Lu, Y., Zeng, F., 2025. A cnn-transformer fusion network for diabetic retinopathy image classification. Computerized Medical Imaging and Graphics , 102655.

Imran, A., Li, J., Pei, Y., Yang, J.J., Wang, Q., 2019. Comparative analysis of vessel segmentation techniques in retinal images. IEEE Access 7, 114862–114887.

International Diabetes Federation (IDF), 2024a. Diabetes facts & figures.

International Diabetes Federation (IDF), 2024b. Idf diabetes atlas: Global diabetes data & statistics.

Karthik, Maggie, Dane, S., 2019. Aptos 2019 blindness detection. https://kaggle.com/competitions/aptos2019-blindness-detection. Kaggle.

Kocak, B., Klontzas, M.E., Stanzione, A., Meddeb, A., Demircioğlu, A., Bluethgen, C., Bressem, K., Ugga, L., Mercaldo, N., Díaz, O., et al., 2025. Evaluation metrics in medical imaging ai: fundamentals, pitfalls, misapplications, and recommendations. European journal of radiology artificial intelligence , 100030.

La Rosa, F., Villari, M., DellAcqua, P., Fazio, M., 2026. Hybrid vision–language models for improved transparency in healthcare processes: the retinal diagnosis use case. Knowledge-Based Systems , 115456.

Li, C., Wong, C., Zhang, S., Usuyama, N., Liu, H., Yang, J., Naumann, T., Poon, H., Gao, J., 2023. Llava-med: Training a large language-and-vision assistant for biomedicine in one day. Advances in Neural Information Processing Systems 36, 28541–28564.

Liu, Z., Lin, Y., Cao, Y., Hu, H., Wei, Y., Zhang, Z., Lin, S., Guo, B., 2021. Swin transformer: Hierarchical vision transformer using shifted windows, in: Proceedings of the IEEE/CVF international conference on computer vision, pp. 10012–10022.

Liu, Z., Mao, H., Wu, C.Y., Feichtenhofer, C., Darrell, T., Xie, S., 2022. A convnet for the 2020s, in: Proceedings of the IEEE/CVF conference on computer vision and pattern recognition, pp. 11976–11986.

Mohsen, F., Belhaouari, S., Shah, Z., 2025. Integrating non-linear radon transformation for diabetic retinopathy grading. Scientific Reports 15, 30706.

Rai, B.B., Maddess, T., Nolan, C.J., 2025. Functional diabetic retinopathy: A new concept to improve management of diabetic retinal diseases. Survey of Ophthalmology 70, 232–240.

Selvaraju, R.R., Cogswell, M., Das, A., Vedantam, R., Parikh, D., Batra, D., 2017. Grad-cam: Visual explanations from deep networks via gradient-based localization, in: Proceedings of the IEEE international conference on computer vision, pp. 618–626.

Sun, L., Peng, Q., Xiao, X., Liu, J., Li, Y., Shu, D., Xue, J., 2025. Mafnet: A novel adaptive multi-scale model for fine-grained grading of diabetic retinopathy. Scientific Reports 15, 32280.

Tan, M., Le, Q., 2019. Efficientnet: Rethinking model scaling for convolutional neural networks, in: International conference on machine learning, PMLR. pp. 6105–6114.

This, M., Tanaka, K., Matsuda, K., Nakayama, A., 2025. Xdr-lvlm: An explainable vision-language large model for diabetic retinopathy diagnosis. arXiv preprint arXiv:2508.15168 .

Tibshirani, R.J., Efron, B., 1993. An introduction to the bootstrap. Monographs on statistics and applied probability 57, 1–436.

Touvron, H., Cord, M., Douze, M., Massa, F., Sablayrolles, A., Jégou, H., 2021. Training data-efficient image transformers & distillation through attention, in: International conference on machine learning, PMLR. pp. 10347–10357.

Wei, Q., Qian, K., Li, X., 2025. Funbench: Benchmarking fundus reading skills of mllms, in: International Conference on Medical Image Computing and Computer-Assisted Intervention, Springer. pp. 278–288.

Wilcoxon, F., 1992. Individual comparisons by ranking methods, in: Breakthroughs in statistics: Methodology and distribution. Springer, pp. 196–202.

World Health Organization Regional Office for Europe, 2026. Promoting diabetic retinopathy screening.

World Health Organization (WHO), 2026. Blindness and vision impairment.

Xu, H., Shao, X., Fang, D., Huang, F., 2024. A hybrid neural network approach for classifying diabetic retinopathy subtypes. Frontiers in Medicine 10, 1293019.

Zang, F., Ma, H., 2024. Cra-net: Transformer guided category-relation attention network for diabetic retinopathy grading. Computers in Biology and Medicine 170, 107993.

Zhang, T., Kishore, V., Wu, F., Weinberger, K.Q., Artzi, Y., 2020. Bertscore: Evaluating text generation with bert, in: International Conference on Learning Representations.

Zhang, W., Belcheva, V., Ermakova, T., 2025a. Interpretable deep learning for diabetic retinopathy: A comparative study of cnn, vit, and hybrid architectures. Computers 14, 187.

Zhang, Z., Zhao, H., Dong, L., Luo, L., Wang, H., 2025b. A study on the interpretability of diabetic retinopathy diagnostic models. Bioengineering 12, 1231.
Khokhar et al.: *Preprint submitted to Elsevier*  Page 20 of 20